\documentclass[a4paper,french]{rnti}  

\usepackage[T1]{fontenc}
\usepackage[utf8]{inputenc}
\usepackage{amsfonts}
\usepackage{amsmath}
\usepackage{hyperref}

\usepackage{url}

\usepackage{graphicx}
\usepackage{multirow}

\titrecourt{Détection d'anomalies par partitionnement}

\nomcourt{P. Lotte et al.}

\titre{Détection d'anomalies par partitionnement des séries temporelles multi-variées}

\auteur{Pierre LOTTE\affil{1},
        André PÉNINOU\affil{2},
        Olivier TESTE\affil{2}}

\affiliation{
    \affil{1}IRIT, Université Toulouse 3 - Paul Sabatier, CNRS\\
          118 Route de Narbonne - 31062 Toulouse, France\\
          pierre.lotte@irit.fr,\\
    \affil{2}IRIT, Université Toulouse 2 - Jean Jaurès, CNRS\\
          118 Route de Narbonne - 31062 Toulouse, France\\
          \{andre.peninou, olivier.teste\}@irit.fr,\\
 }

\resume{
  Dans cet article, nous proposons une méthode non-supervisée de détection des anomalies dans les séries temporelles multi-variées par partitionnement appelée PARADISE. Cette méthode consiste à créer une partition des variables de la série temporelle étudiée en garantissant la conservation des relations inter-variables servant à identifier les anomalies. Ce partitionnement repose sur un clustering de plusieurs corrélations entre variables. La détection des anomalies est réalisée localement sur chacune des parties. Plusieurs expérimentations effectuées sur des jeux de données synthétiques et réels issus de la littérature, montrent la pertinence de l'approche avec une amélioration significative de la détection d’anomalies.
}

\summary{
  In this article, we suggest a novel non-supervised partition based anomaly detection method for anomaly detection in multivariate time series called PARADISE. This methodology creates a partition of the variables of the time series while ensuring that the inter-variable relations remain untouched. This partitioning relies on the clustering of multiple correlation coefficients between variables to identify subsets of variables before executing anomaly detection algorithms locally for each of those subsets. Through multiple experimentations done on both synthetic and real datasets coming from the literature, we show the relevance of our approach with a significant improvement in anomaly detection performance.
}

\graphicspath{{./assets/}}

\begin{document}

%
\section{Introduction}

Les séries temporelles sont présentes dans de nombreux domaines tels que la santé, la finance, la sécurité des systèmes, l'aéronautique, l'Internet des objets (IOT), les bâtiments connectés (Smart Building) et bien d'autres. Une série temporelle est constituée d'une ou plusieurs variables mesurées au cours du temps. On qualifie une série temporelle d'\textit{uni-variée} ou de \textit{multi-variée} lorsque celle-ci est composée respectivement d'une ou plusieurs variables.

Une des tâches les plus importantes concernant les séries temporelles est la détection d'anomalies qui consiste à identifier les valeurs aberrantes de la série temporelle étudiée. Plusieurs approches ont été proposées dans la littérature scientifique~\cite{schmidlAnomalyDetectionTime2022}.

Les séries temporelles multi-variées sont des ensembles de données complexes constitués de nombreuses variables entre lesquelles peuvent exister des relations inter-variables importantes mais difficiles à exploiter~\cite{lejeuneShapebasedOutlierDetection2020}. La majorité des approches existantes de détection d'anomalies exploitent les relations temporelles et inter-variables sur l'ensemble des variables qui constituent les séries temporelles multi-variées, sans identifier de sous-ensembles de variables pouvant modéliser localement des phénomènes reliés entre eux, et relativement indépendants des autres variables. La connaissance de ces sous-ensembles de variables pourrait permettre d'identifier et d'exploiter plus facilement les relations inter-variables.

Nous définissons dans cet article une méthode de détection d'anomalies non supervisée par partitionnement de l'ensemble des variables dans les séries temporelles multi-variées. Notre méthode appelée PARADISE (\textbf{PAR}tition-based \textbf{A}nomaly \textbf{D}etection for multivariate t\textbf{I}me \textbf{SE}ries) identifie dans un premier temps les sous-ensembles de variables fortement corrélées entre elles et faiblement corrélées avec les autres, puis dans un deuxième temps, effectue une détection d'anomalies locale à chaque sous-ensemble (et non à l'ensemble des variables). Cet article vise à montrer l'intérêt de cette démarche sur des jeux de données synthétiques et réels issus de la littérature scientifique.

Nous commençons cet article par un état de l'art de la littérature scientifique portant sur la détection d'anomalies dans les séries temporelles multi-variées. Nous détaillons ensuite la méthode PARADISE proposée. Enfin, nous discutons des expérimentations effectuées dans le but de montrer la pertinence d'une telle approche.

\section{État de l'art}

Parmi les premières solutions automatiques proposées pour la détection d'anomalies dans les séries temporelles multi-variées, la majorité se base sur l'utilisation de modèles statistiques tels que ARIMA proposé dans ~\cite{boxTimeSeriesAnalysis2015} ou encore de l'une de ses variantes. Par la suite, les chercheurs se sont penchés sur l'utilisation d'algorithmes d'apprentissage machine. ~\cite{breunigLOFIdentifyingDensitybased2000} proposent par exemple un nouvel algorithme appelé LOF (Local Outlier Factor). Cette méthode se base sur la densité du voisinage de chaque échantillon d'une série temporelle pour en déterminer un degré d'anomalie.~\cite{yairiFaultDetectionMining2001} utilisent l'algorithme K-Means afin de réaliser un clustering permettant d'identifier les motifs récurrents qui peuvent ensuite être liés par des règles d'associations décrivant le comportement normal du système.~\cite{liuIsolationBasedAnomalyDetection2012} proposent l'algorithme Isolation Forest (IF) qui utilise des arbres binaires dans lesquels on tente d'isoler chaque échantillon de la série temporelle à travers un ensemble d'attributs. Pour aller encore plus loin,~\cite{chengOutlierDetectionUsing2019} propose de combiner LOF et IF.

Depuis quelques années, la majeure partie des solutions proposées se basent sur des techniques d'apprentissage profond. Les solutions proposées dans~\cite{malhotraLSTMbasedEncoderDecoderMultisensor2016},~\cite{suRobustAnomalyDetection2019},~\cite{munirDeepAnTDeepLearning2019},~\cite{chenImbalancedDatasetbasedEcho2020} utilisent une grande variété de modèles tels que des auto-encodeurs basés sur des cellules LSTM, des réseaux de neurones à convolution ou des ``echo state networks'' (ESN) \footnote{ESN: Réseaux de neurones tentant de se rapprocher du modèle d'un cerveau humain en se basant sur des techniques de \textit{reservoir computing}} par exemple. Ces articles se basent tous sur le même principe. Leur entraînement est réalisé uniquement à partir de données ne comportant aucune anomalie. Lorsqu'on leur demande de prédire le futur ou de reconstruire le signal fourni en entrée, ces modèles se retrouveront alors loin de la vérité lorsque les données contiennent des anomalies. Bien que performantes, aucune de ces approches ne prend en compte l'existence de sous-ensembles de variables contenant des relations inter-variables intéressantes.

Dans les approches par apprentissage profond on peut distinguer les méthodes utilisant des réseaux de neurones à graphes (GNN) comme celles de~\cite{dengGraphNeuralNetworkBased2021} ou encore de~\cite{dingMSTGATMultimodalSpatial2023}. Dans ces approches, on modélise les données sous forme de graphes dans lesquels chaque noeud représente une variable du jeu de données et chaque arête représente une relation inter-variable entre deux noeuds. Bien que cette modélisation induise une séparation des variables, la façon dont cette séparation est réalisée ne garantit pas la conservation des relations inter-variables présentes dans le jeu de données utilisé ce qui pourrait provoquer une perte d'informations.

Malgré les gains en performance obtenus par chacune des générations de solution proposées, la détection d'anomalies dans les jeux de données à grande dimensionalité présents dans la littérature scientifique reste difficile pour les algorithmes. Nous suspectons que cette difficulté soit liée à la malédiction de la dimensionalité et à l'absence d'identification de sous-ensembles de variables liées.

\section{Description de l'approche PARADISE}

\subsection{Formalisation du problème}

Soit une série temporelle multi-variée notée $X$ composée de $d$ variables et $n$ observations définie par $X = \begin{pmatrix}x_{i,j}\end{pmatrix}_{1\leq i\leq n,\\ 1\leq j \leq d}, x_{i,j}\in\mathbb{R}$. On notera $x_{i,\cdot}$ toute observation de $X$ pour l'instant $i$ et $x_{\cdot, j}$ toute variable $j$ de $X$.

On cherche à affecter à chaque observation $x_{i, \cdot}$ de $X$ un label noté $y_{i}$ afin d'obtenir le vecteur $Y=(y_{i}), y_{i}\in\{0,1\}, 1\leq i\leq n$. Les observations considérées comme anormales seront marquées $1$ et les observations normales sont marquées $0$.

L'approche PARADISE crée une partition de $X$ notée $\mathcal{P}$. Chaque partie notée $X^{k}= \{x_{\cdot, j_{1}}, x_{\cdot, j_{2}},\dots\}$ est définie comme un ensemble non vide de variables contenues dans $X$. La partition $\mathcal{P}$ est telle que $X^{1}\cup X^{2}\cup\dots\cup X^{p} = X$ et pour toutes parties $X^k$ et $X^l$ différentes, $X^k\cap X^l=\emptyset$.

On modélise les relations inter-variables par des fonctions notées $f$. Soient $j_{1}\in \{1..d\}$ et $j_{2}\in \{1..d\}$ une relation inter-variable existe entre $x_{\cdot,j_{1}}$ et $x_{\cdot,j_{2}}$ si $\exists f: \mathbb{R}^{n} \rightarrow \mathbb{R}^{n}, f(x_{\cdot, j_{1}}) = x_{\cdot,j_{2}}$.

On note $\mathcal{F}$ l'ensemble des relations inter-variables, la partition idéale notée $\mathcal{P}^{*}$ est une partition qui conserve l'ensemble $\mathcal{F}$. Elle respecte les règles suivantes: (i) $\forall f \in \mathcal{F}, \exists X^{k}\in\mathcal{P} | f(x_{\cdot,j_{1}}) = x_{\cdot,j_{2}} \land x_{\cdot,j_{1}} \in X^{k} \land x_{\cdot,j_{2}} \in X^{k}$, (ii) Soit l'opérateur $ x_{\cdot, j_1} \rightarrow x_{\cdot, j_2}$ qui signifie $\exists f\in\mathcal{F} | f(x_{\cdot, j_1}) = x_{\cdot, j_2} \lor f(x_{\cdot, j_2}) = x_{\cdot, j_1}$. On a $\forall x_{\cdot,j_1} \in X^k, \forall x_{\cdot,j_2} \in X^k, \exists \{x_{\cdot,l_1},\dots,x_{\cdot,l_m}\} \subset X^k | x_{\cdot,j_1}\rightarrow x_{\cdot, l_1}\rightarrow \dots\rightarrow x_{\cdot, l_m}\rightarrow x_{\cdot, j_2}$

\subsection{Partitionnement}\label{sec:clustering}

La première étape mise en place dans PARADISE est une étape de séparation des variables par création d'une partition notée $\widehat{\mathcal{P}^{*}}$ qui se veut être la plus proche de $\mathcal{P}^*$ ($\mathcal{P}^*$ n'est pas connu dans les jeux de données réels). Pour créer cette partition, nous identifions les relations inter-variables dans les données sans connaissances préalables du système étudié. Pour cela nous utilisons des coefficients de corrélation dans le but de mesurer le degré de relation entre deux variables. D'autres approches de recherche des partitions pourront par la suite être étudiées.

Il existe différents coefficients de corrélation possédant chacun des capacités de détection différentes. Par exemple, le coefficient de Pearson, détecte seulement les relations linéaires. En revanche d'autres coefficients comme ceux de Kendall, Spearman, la corrélation de distance~\cite{szekelyMeasuringTestingDependence2007} ou le coefficient $\xi$~\cite{chatterjeeNewCoefficientCorrelation2021} peuvent détecter certaines relations non-linéaires. Pour favoriser la robustesse de notre approche de détection des relations inter-variables, nous utiliserons ces cinq coefficients. La valeur retenue pour chaque couple de variables est la valeur absolue maximale atteint par un des cinq coefficients.

Nous allons nous baser sur ces coefficients de corrélation entre variables pour déterminer les sous-ensembles de variables possédant des relations inter-variables pertinentes. Pour trouver ces sous-ensembles, notre méthode s'appuie sur l'utilisation d'algorithmes de clustering tels que K-Means ou HDBSCAN. En effet, chaque ligne de la matrice de corrélation donne les coordonnées d'un point dans un espace multi-dimensionnel $(d\times d)$, nous considérons que les points faisant parti d'un sous-ensemble de variables liées seront proches les uns des autres et isolés du reste des points. L'algorithme de clustering utilisé pourra alors identifier ces sous-ensembles.

En utilisant ce principe, PARADISE est capable de proposer une partition $\widehat{\mathcal{P}^{*}}$ conservant les relations inter-variables au maximum.

\subsection{Détection des anomalies par partie}\label{sec:exec}

Une fois la partition obtenue, nous pouvons entraîner et exécuter les algorithmes de détection d'anomalies localement sur chacune des parties de manière indépendante. Les algorithmes utilisés acceptent en entrée  toute partie $X^{k} \in \widehat{\mathcal{P}^{*}}$ préalablement créée et fourniront en sortie un ensemble de scores d'anomalie noté $S^k=\{s^k_1, s^k_2, \dots, s^k_n\}$ où le score obtenu par chaque observation $x_{i, \cdot}$ de la partie $X^k$ sera noté $s^k_i$.

À partir des scores locaux affectés aux différentes parties nous devons calculer un score global pour la série temporelle multivariée. Pour ce faire, nous normalisons entre $0$ et $1$ les scores d'anomalies obtenus par partie de manière indépendante. Cette normalisation a pour but de rendre les scores locaux comparables en terme d'échelle. Ensuite, et puisque nous sommes intéressés principalement par les scores élevés, nous conserverons pour chaque instant de la série temporelle, le score d'anomalie maximal obtenu par toutes les parties.

Puisque nous passons par une première étape de détection locale, nous obtenons, en plus du score d'anomalie global, une indication plus précise sur l'origine de chaque anomalie. En effet, nous savons quelle partie est à l'origine du score de chacune des observations et pouvont donc réduire le nombre de variables candidates ce qui permet d'expliquer en partie l'anomalie.

\section{Expérimentations}

\subsection{Protocole expérimental}

\paragraph{Jeux de données synthétiques} Nous utilisons des jeux de données synthétiques générés par un outil développé dans le cadre de nos travaux\footnote{\href{https://gitlab.irit.fr/sig/theses/pierre-lotte/PARADISE}{https://gitlab.irit.fr/sig/theses/pierre-lotte/PARADISE}}. Les jeux de données générés contiennent un nombre de variables, un taux de contamination et un nombre de sous-ensembles de variables liées configurables. Chaque variable peut être considérée comme une variable de support ou de suivi. Les variables de support sont représentées par des combinaisons linéaires de fonctions sinusoïdales. Leurs équations sont donc décrites par la forme générale $f(x) = \sum^{m}_{i=1}\beta_{i}osc_{i}(\alpha_{i}x)$ avec $\beta_{i}$ l'amplitude et $\alpha_{i}$ la fréquence de la $i^{\text{ème}}$ fonction sinusoïdale $osc_{i}$. La fréquence peut également être variable plutôt que fixe.

Les variables de suivi sont quant à elles des variables corrélées aux variables de support. Ces variables peuvent être corrélées de manière linéaire, exponentielle ou logarithmique. Leur génération se base sur un système de suites. Si on prend le cas d'une variable de suivi basée sur une corrélation linéaire, les points de la série temporelle générée sont les termes successifs de la suite $u_{m+1} = u_{m} + sgn\left(f(m+1)-f(m)\right)\times r$ avec $r$ le pas effectué entre chaque point successif, $sgn$ la fonction signe et $f$ la fonction de support. Le principe de construction est identique pour les variables corrélées de façon exponentielle ou logarithmique.

Une fois les données générées, l'outil injecte des anomalies dans certaines variables de chaque sous-ensemble. Les anomalies peuvent être des anomalies de bruit, de fréquence ou encore de suivi des corrélations. Lors de l'injection d'une anomalie, on ne modifie les valeurs que de la variable affectée.

\paragraph{Jeux de données réels} Nous utilisons également des jeux de données réels couramment utilisés dans la littérature scientifique. Le premier jeu de données, nommé \textbf{SMD}, publié dans~\cite{suRobustAnomalyDetection2019} contient les relevés de 30 capteurs pour 28 serveurs d'un datacenter. Le deuxième nommé \textbf{WADI}, publié dans~\cite{ahmedWADIWaterDistribution2017}, provient d'un système de distribution de l'eau contenant 93 capteurs. Enfin, le troisième nommé \textbf{SWaT}, publié dans~\cite{mathurSWaTWaterTreatment2016}, provient d'un système de traitement de l'eau contenant 44 capteurs. Les taux de contaminations de ces jeux de données sont respectivement de $4.21\%$, $5.77\%$ et $17.37\%$.

\paragraph{Algorithmes de référence utilisés} Nous vérifions l'efficacité de notre méthode avec divers algorithmes. Nous avons retenu IForest (\cite{liuIsolationBasedAnomalyDetection2012}), LOF (\cite{breunigLOFIdentifyingDensitybased2000}), K-Means (\cite{yairiFaultDetectionMining2001}), DeepANT (\cite{munirDeepAnTDeepLearning2019}) et HealthESN (\cite{chenImbalancedDatasetbasedEcho2020}). Le choix de ces algorithmes s'est fait sur deux critères: (i) avoir de la diversité dans les approches utilisées, (ii) retenir les algorithmes parmi ceux ayant le mieux performé dans les expérimentations menées par~\cite{schmidlAnomalyDetectionTime2022}.

\subsection{Résultats}

\paragraph{QR1: L'approche PARADISE permet-elle d'améliorer les performances de détection d'anomalies ?} Les premières expérimentations réalisées portent sur les jeux de données synthétiques pour lesquels la partition idéale est connue par construction. Les résultats obtenus lors de ces expérimentations sont détaillés dans la table~\ref{tab:results_synth}. Ces expérimentations ont été menées sur $132$ jeux de données comportant entre $5$ et $50$ variables, $20$k observations, entre $2$ et $20$ parties et à des taux de contamination allant de $0.1\%$ à $10\%$. Pour effectuer nos expérimentations, nous utiliserons les quatre métriques suivantes: le F1 score (F1), la précision (Pr), le rappel (Ra) et le score ROC (ROC).  Le F1 score est obtenu par optimisation du seuil de détection des anomalies grâce à la courbe ROC. Pour PARADISE, le partitionnement obtenu est le meilleur trouvé par clustering avec K-Means et HDBSCAN, optimisé sur le ROC final.

\begin{table}[h!]
  \centering
  \begin{tabular}{|c|cccc|cccc||cccc|}
    \hline
    & \multicolumn{4}{|c|}{App. Classique} & \multicolumn{4}{|c||}{PARADISE} & \multicolumn{4}{|c|}{Part. idéal $P^{*}$} \\
    \hline
    Algorithme & F1 & Pr & Ra & ROC & F1 & Pr & Ra & ROC & F1 & Pr & Ra & ROC \\
    \hline
    DeepANT & $\underline{.12}$ & $.07$ & $\underline{.63}$ & $.52$ & $\underline{.12}$ & $\underline{.08}$ & $\underline{.63}$ & $\underline{.53}$ & $\textbf{.15}$ & $\textbf{.10}$ & $\textbf{.65}$ & $\textbf{.60}$ \\
    HealthESN & $.11$ & $.07$ & $.57$ & $.57$ & $\textbf{.15}$ & $\textbf{.10}$ & $\textbf{.63}$ & $\underline{.66}$ & $\textbf{.15}$ & $\textbf{.10}$ & $\underline{.62}$ & $\textbf{.67}$ \\
    IForest & $.10$ & $.06$ & $\textbf{.57}$ & $.51$ & $\textbf{.11}$ & $\textbf{.07}$ & $\underline{.56}$ & $\textbf{.52}$ & $\textbf{.11}$ & $\textbf{.07}$ & $.54$ & $\textbf{.52}$ \\
    K-Means & $.10$ & $.06$ & $.55$ & $.51$ & $\underline{.11}$ & $\underline{.07}$ & $\underline{.57}$ & $\underline{.56}$ & $\textbf{.13}$ & $\textbf{.08}$ & $\textbf{.58}$ & $\textbf{.58}$ \\
    LOF & $.10$ & $.06$ & $.52$ & $.50$ & $\textbf{.11}$ & $\textbf{.07}$ & $\textbf{.53}$ & $\textbf{.53}$ & $\textbf{.11}$ & $\textbf{.07}$ & $\textbf{.53}$ & $\textbf{.53}$ \\
    \hline
  \end{tabular}
  \caption{Résultats obtenus sur les jeux de données synthétiques. Pour les deux métriques, la version la plus performante est en gras et la deuxième est soulignée.\label{tab:results_synth}}
\end{table}

De ces résultats nous pouvons conclure que notre approche permet un gain de performance significatif dans toutes les métriques dans une grande majorité de cas. Le gain obtenu par notre approche dépend des algorithmes utilisés. Certains algorithmes comme HealthESN, DeepANT et K-Means obtiennent des gains moyens de $10\%$, $8\%$ et $7\%$ respectivement entre l'approche classique et le partitionnement idéal pour la métrique ROC. En revanche, les algorithmes basés sur IForest et LOF bénéficient moins du partitionnement. Nous pouvons également observer que la partition proposée par PARADISE n'atteint pas toujours les résultats obtenus par la partition idéale mais dépasse les résultats de l'approche classique. Le partitionnement proposé ne doit donc pas conserver toutes les relations inter-variables. Ceci est sans doute dû au fait que les coefficients de corrélation, bien que pertinents, sont des outils insuffisants.

Nous avons ensuite procédé à l'exécution des mêmes expérimentations sur les jeux de données issus de la littérature scientifique. Les résultats obtenus sont décrits dans la table~\ref{tab:results_reels}.

\begin{table}[h!]
  \centering
  \begin{tabular}{|c|c|cccc|cccc|}
    \hline
    & Algorithme & \multicolumn{4}{|c|}{Classique} & \multicolumn{4}{|c|}{PARADISE}\\
    \hline
    & & F1 & Pr & Ra & ROC & F1 & Pr & Ra & ROC \\
    \hline
    \multirow{4}{*}{\rotatebox{90}{WADI}} & \multicolumn{1}{|c|}{DeepANT} & $\textbf{.16}$ & $\textbf{.09}$ & $\textbf{.54}$ & $\textbf{.61}$ & $.12$ & $.06$ & $\textbf{.54}$ & $.56$ \\
    & \multicolumn{1}{|c|}{HealthESN} & $\textbf{.16}$ & $\textbf{.10}$ & $.65$ & $.42$ & $.12$ & $.06$ & $\textbf{.68}$ & $\textbf{.63}$ \\
    & \multicolumn{1}{|c|}{IForest} & $\textbf{.25}$ & $\textbf{.15}$ & $\textbf{.69}$ & $\textbf{.77}$ & $.24$ & $.14$ & $\textbf{.69}$ & $.74$ \\
    & \multicolumn{1}{|c|}{K-Means} & $.10$ & $.06$ & $.38$ & $.48$ & $\textbf{.19}$ & $\textbf{.11}$ & $\textbf{.72}$ & $\textbf{.72}$ \\
    & \multicolumn{1}{|c|}{LOF} & $\textbf{.11}$ & $\textbf{.06}$ & $\textbf{.51}$ & $\textbf{.51}$ & $.10$ & $\textbf{.06}$ & $.48$ & $.50$ \\
    \hline
    \multirow{4}{*}{\rotatebox{90}{SWaT}} & \multicolumn{1}{|c|}{DeepANT} & $\textbf{.34}$ & $\textbf{.26}$ & $\textbf{.52}$ & $.41$ & $.32$ & $\textbf{.26}$ & $.43$ & $\textbf{.45}$ \\
    & \multicolumn{1}{|c|}{HealthESN} & $\textbf{.56}$ & $\textbf{.47}$ & $.68$ & $\textbf{.60}$ & $.43$ & $.30$ & $\textbf{.71}$ & $.46$ \\
    & \multicolumn{1}{|c|}{IForest} & $.28$ & $.22$ & $.37$ & $.35$ & $\textbf{.30}$ & $\textbf{.23}$ & $\textbf{.45}$ & $\textbf{.36}$ \\
    & \multicolumn{1}{|c|}{K-Means} & $\textbf{.37}$ & $\textbf{.28}$ & $\textbf{.52}$ & $\textbf{.45}$ & $.27$ & $.21$ & $.35$ & $.38$ \\
    & \multicolumn{1}{|c|}{LOF} & $\textbf{.36}$ & $\textbf{.29}$ & $.49$ & $\textbf{.49}$ & $\textbf{.36}$ & $.28$ & $.51$ & $\textbf{.49}$ \\
    \hline
    \multirow{4}{*}{\rotatebox{90}{SMD}} & \multicolumn{1}{|c|}{DeepANT} & $\textbf{.25}$ & $\textbf{.17}$ & $.73$ & $.74$ & $.22$ & $.15$ & $\textbf{.74}$ & $\textbf{.75}$\\
    & \multicolumn{1}{|c|}{HealthESN} & $\textbf{.25}$ & $\textbf{.18}$ & $\textbf{.77}$ & $\textbf{.83}$ & $.23$ & $.15$ & $.76$ & $.82$ \\
    & \multicolumn{1}{|c|}{IForest} & $\textbf{.23}$ & $\textbf{.14}$ & $\textbf{.76}$ & $\textbf{.83}$ & $\textbf{.23}$ & $\textbf{.14}$ & $.75$ & $.82$ \\
    & \multicolumn{1}{|c|}{K-Means} & $.29$ & $.19$ & $\textbf{.81}$ & $\textbf{.87}$ & $\textbf{.30}$ & $\textbf{.20}$ & $\textbf{.81}$ & $\textbf{.87}$ \\
    & \multicolumn{1}{|c|}{LOF} & $\textbf{.11}$ & $\textbf{.14}$ & $.58$ & $\textbf{.67}$ & $\textbf{.11}$ & $\textbf{.14}$ & $\textbf{.59}$ & $\textbf{.67}$ \\
    \hline
  \end{tabular}
  \caption{Résultats obtenus sur les jeux de données réels. L'approche la plus performante pour chaque métrique est en gras.\label{tab:results_reels}}
\end{table}

Dans cette table, nous pouvons remarquer plusieurs choses. Tout d'abord, l'efficacité de l'approche PARADISE dépend de l'algorithme mais aussi du jeu de données. Par exemple, l'algorithme HealthESN obtient une amélioration importante de ses performances sur WADI mais perd sur SWaT et reste stable sur SMD. Toutefois, l'approche PARADISE n'induit de perte de performance significative que rarement. Les performances atteintes sur les jeux de données réels ne confirment pas totalement les conclusions précédentes. Cela est probablement dû à la complexité plus importante des jeux de données réels pour lesquels les relations inter-variables sont plus difficiles à détecter. Dans la suite de nos expérimentations, nous écarterons les algorithmes peu sensibles à notre méthode à savoir IForest et LOF.

\paragraph{QR2: Est-ce que le partitionnement proposé conserve au maximum les relations inter-variables ?} L'approche PARADISE doit conserver les relations inter-variables présentes dans les données afin de conserver les informations. Pour vérifier cela, nous avons utilisé la métrique \textbf{ARI}. Nous avons étudié les valeurs de cette métrique pour les partitionnements proposés par notre méthode sur les jeux de données de plus de $30$ variables comportant entre $4$ et $20$ sous-ensembles de variables.

Bien que pertinent dans de nombreux cas, notre partitionnement n'est pas assez précis. En effet, la majorité des cas de figures nous donne un score ARI aux alentours de $0.45$ avec des pics pouvant aller jusqu'à $0.72$. Les jeux de données possédant peu de sous-ensembles de variables semblent être plus difficiles à partitionner, ce qui se traduit par des scores faibles se situant autour de $0.10$. Cela explique les différences obtenues par les partitionnements proposé et idéal observées dans la table~\ref{tab:results_synth}.

\section{Conclusion}

Nous décrivons dans cet article une nouvelle approche de détection des anomalies par partitionnement dans les séries temporelles multi-variées. Cette approche partitionne les variables des séries temporelles en utilisant des coefficients de corrélation et des algorithmes de clustering. Les parties sont ensuite traitées séparément par les algorithmes de détection d'anomalies qui leur affectent des scores locaux. Ces scores locaux sont utilisés pour calculer un score global pour la série temporelle multi-variée. Après plusieurs expérimentations menées sur des jeux de données synthétiques et issus de la littérature, nous montrons que notre approche permet d'obtenir des gains de performances souvent significatifs peu importe le nombre de parties. Dans la suite de nos travaux nous nous pencherons sur l'amélioration de la méthode de partitionnement afin de détecter plus finement les relations entre variables.

\bibliographystyle{rnti}
\bibliography{biblio}

\providecommand\Fr{}
\providecommand\Eng{}
\providecommand\andname{and}
\providecommand\andnamec{and}

\begin{thebibliography}{}


\bibitem[{Ahmed et~al.}(2017){Ahmed, Palleti, \andnamec{}
  Mathur}]{ahmedWADIWaterDistribution2017}
Ahmed, C.~M., V.~R. Palleti, \andname{} A.~P. Mathur (2017).
\newblock {{WADI}}: A water distribution testbed for research in the design of
  secure cyber physical systems.
\newblock In {\em Proceedings of the 3rd {{International Workshop}} on
  {{Cyber-Physical Systems}} for {{Smart Water Networks}}}, {{CySWATER}} '17,
  New York, NY, USA, pp.\  25--28. Association for Computing Machinery.

\bibitem[{Box et~al.}(2015){Box, Jenkins, Reinsel, \andnamec{}
  Ljung}]{boxTimeSeriesAnalysis2015}
Box, G. E.~P., G.~M. Jenkins, G.~C. Reinsel, \andname{} G.~M. Ljung (2015).
\newblock {\em Time {{Series Analysis}}: {{Forecasting}} and {{Control}}}.
\newblock John Wiley \& Sons.

\bibitem[{Breunig et~al.}(2000){Breunig, Kriegel, Ng, \andnamec{}
  Sander}]{breunigLOFIdentifyingDensitybased2000}
Breunig, M.~M., H.-P. Kriegel, R.~T. Ng, \andname{} J.~Sander (2000).
\newblock {{LOF}}: Identifying density-based local outliers.
\newblock In {\em Proceedings of the 2000 {{ACM SIGMOD}} International
  Conference on {{Management}} of Data}, Dallas Texas USA, pp.\  93--104. ACM.

\bibitem[{Chatterjee}(2021){Chatterjee}]{chatterjeeNewCoefficientCorrelation2021}
Chatterjee, S. (2021).
\newblock A {{New Coefficient}} of {{Correlation}}.
\newblock {\em Journal of the American Statistical Association\/}~{\em
  116\/}(536), 2009--2022.

\bibitem[{Chen et~al.}(2020){Chen, Zhang, Huang, He, \andnamec{}
  Song}]{chenImbalancedDatasetbasedEcho2020}
Chen, Q., A.~Zhang, T.~Huang, Q.~He, \andname{} Y.~Song (2020).
\newblock Imbalanced dataset-based echo state networks for anomaly detection.
\newblock {\em Neural Computing and Applications\/}~{\em 32\/}(8), 3685--3694.

\bibitem[{Cheng et~al.}(2019){Cheng, Zou, \andnamec{}
  Dong}]{chengOutlierDetectionUsing2019}
Cheng, Z., C.~Zou, \andname{} J.~Dong (2019).
\newblock Outlier detection using isolation forest and local outlier factor.
\newblock In {\em Proceedings of the {{Conference}} on {{Research}} in
  {{Adaptive}} and {{Convergent Systems}}}, {{RACS}} '19, New York, NY, USA,
  pp.\  161--168. Association for Computing Machinery.

\bibitem[{Deng \andnamec{} Hooi}(2021){Deng \andnamec{}
  Hooi}]{dengGraphNeuralNetworkBased2021}
Deng, A. \andname{} B.~Hooi (2021).
\newblock Graph {{Neural Network-Based Anomaly Detection}} in {{Multivariate
  Time Series}}.
\newblock {\em Proceedings of the AAAI Conference on Artificial
  Intelligence\/}~{\em 35\/}(5), 4027--4035.

\bibitem[{Ding et~al.}(2023){Ding, Sun, \andnamec{}
  Zhao}]{dingMSTGATMultimodalSpatial2023}
Ding, C., S.~Sun, \andname{} J.~Zhao (2023).
\newblock {{MST-GAT}}: {{A}} multimodal spatial--temporal graph attention
  network for time series anomaly detection.
\newblock {\em Information Fusion\/}~{\em 89}, 527--536.

\bibitem[{Lejeune et~al.}(2020){Lejeune, Mothe, Soubki, \andnamec{}
  Teste}]{lejeuneShapebasedOutlierDetection2020}
Lejeune, C., J.~Mothe, A.~Soubki, \andname{} O.~Teste (2020).
\newblock Shape-based outlier detection in multivariate functional data.
\newblock {\em Knowledge-Based Systems\/}~{\em 198}, 105960.

\bibitem[{Liu et~al.}(2012){Liu, Ting, \andnamec{}
  Zhou}]{liuIsolationBasedAnomalyDetection2012}
Liu, F.~T., K.~M. Ting, \andname{} Z.-H. Zhou (2012).
\newblock Isolation-{{Based Anomaly Detection}}.
\newblock {\em ACM Trans. Knowl. Discov. Data\/}~{\em 6\/}(1), 3:1--3:39.

\bibitem[{Malhotra et~al.}(2016){Malhotra, Ramakrishnan, Anand, Vig, Agarwal,
  \andnamec{} Shroff}]{malhotraLSTMbasedEncoderDecoderMultisensor2016}
Malhotra, P., A.~Ramakrishnan, G.~Anand, L.~Vig, P.~Agarwal, \andname{}
  G.~Shroff (2016).
\newblock {{LSTM-based Encoder-Decoder}} for {{Multi-sensor Anomaly
  Detection}}.

\bibitem[{Mathur \andnamec{} Tippenhauer}(2016){Mathur \andnamec{}
  Tippenhauer}]{mathurSWaTWaterTreatment2016}
Mathur, A.~P. \andname{} N.~O. Tippenhauer (2016).
\newblock {{SWaT}}: A water treatment testbed for research and training on
  {{ICS}} security.
\newblock In {\em 2016 {{International Workshop}} on {{Cyber-physical Systems}}
  for {{Smart Water Networks}} ({{CySWater}})}, pp.\  31--36.

\bibitem[{Munir et~al.}(2019){Munir, Siddiqui, Dengel, \andnamec{}
  Ahmed}]{munirDeepAnTDeepLearning2019}
Munir, M., S.~A. Siddiqui, A.~Dengel, \andname{} S.~Ahmed (2019).
\newblock {{DeepAnT}}: {{A Deep Learning Approach}} for {{Unsupervised Anomaly
  Detection}} in {{Time Series}}.
\newblock {\em IEEE Access\/}~{\em 7}, 1991--2005.

\bibitem[{Schmidl et~al.}(2022){Schmidl, Wenig, \andnamec{}
  Papenbrock}]{schmidlAnomalyDetectionTime2022}
Schmidl, S., P.~Wenig, \andname{} T.~Papenbrock (2022).
\newblock Anomaly detection in time series: A comprehensive evaluation.
\newblock {\em Proceedings of the VLDB Endowment\/}~{\em 15\/}(9), 1779--1797.

\bibitem[{Su et~al.}(2019){Su, Zhao, Niu, Liu, Sun, \andnamec{}
  Pei}]{suRobustAnomalyDetection2019}
Su, Y., Y.~Zhao, C.~Niu, R.~Liu, W.~Sun, \andname{} D.~Pei (2019).
\newblock Robust {{Anomaly Detection}} for {{Multivariate Time Series}} through
  {{Stochastic Recurrent Neural Network}}.
\newblock In {\em Proceedings of the 25th {{ACM SIGKDD International
  Conference}} on {{Knowledge Discovery}} \& {{Data Mining}}}, {{KDD}} '19, New
  York, NY, USA, pp.\  2828--2837. Association for Computing Machinery.

\bibitem[{Sz{\'e}kely et~al.}(2007){Sz{\'e}kely, Rizzo, \andnamec{}
  Bakirov}]{szekelyMeasuringTestingDependence2007}
Sz{\'e}kely, G.~J., M.~L. Rizzo, \andname{} N.~K. Bakirov (2007).
\newblock Measuring and testing dependence by correlation of distances.
\newblock {\em The Annals of Statistics\/}~{\em 35\/}(6).

\bibitem[{Yairi et~al.}(2001){Yairi, Kato, \andnamec{}
  Hori}]{yairiFaultDetectionMining2001}
Yairi, T., Y.~Kato, \andname{} K.~Hori (2001).
\newblock Fault {{Detection}} by {{Mining Association Rules}} from
  {{House-keeping Data}}.
\newblock ~{\em 3\/}(9).

\end{thebibliography}

\appendix

\Fr

\end{document}